# Optimization of Complex Process, Based on Design Of Experiments, a Generic Methodology

*Julien Baderot[1], Yann Cauchepin[1,2], Alexandre Seiller[1,3], Richard Fontanges[1], Sergio Martinez[1], Johann Foucher[1], Emmanuel Fuchs[4], Mehdi Daanoune[4], Vincent Grenier[4], Vincent Barra[2], Arnaud Guillin[3]*

[1]**Pollen Metrology, Moirans, France**

[2]**UCA-LIMOS, Clermont-Ferrand, France**

[3]**UCA-LMBP, Clermont-Ferrand, France**

[4]**Aledia, Grenoble, France**

**Abstract**

*MicroLED displays are the result of a complex manufacturing chain. Each stage of this process, if optimized, contributes to achieving the highest levels of final efficiencies. Common works carried out by Pollen Metrology, Aledia, and Université Clermont-Auvergne led to a generic process optimization workflow. This software solution offers a holistic approach where stages are chained together for gaining a complete optimal solution. This paper highlights key corners of the methodology, validated by the experiments and process experts: data cleaning and multi-objective optimization.*

## Objective and background

MicroLED processes involve many steps which are difficult to optimize together due to interactions between these different steps. Process engineers must work carefully during experimental optimization to gain knowledge of the process in addition to improving the performance of the device. The cost of experiments, and the number of different Design Of Experiments (DOE), is also a major constraint limiting the quantity of available data to draw conclusions.

Based on the context, process optimization with AI from experimental data raises fundamental questions about the relevance of the data to properly model the process, then the capability of the model to accurately predict multiple output criteria. As an experimental platform, MicroLED with their fabrication recipes, which contains properties such as geometries and physical performances is an ideal testing ground to explore a new AI-based software platform dedicated to process optimization. The proposal is to create a framework for building AI models based on experimental data, by setting appropriate constraints and rules, allowing to predict the most suitable model for an effective optimization, oriented to help process engineers in their work.

Our proposal for the introduction of AI in the process optimization pipeline is to first clean and arrange the data automatically to be compatible with data science algorithms. The second step gathers the training and evaluation of AI models to explain the physical process from data. Finally with these models, a global optimization schema is proposed to explore, without new experiments, the process and provide potential recipes, i.e. guidelines for physical experiments. The standard design of experiments shows that physical intuitions are difficult to establish with multiple modifications of an experiment, whereas AI can process relationships between multiple parameters. These physical explanations from the data can be then confronted by process experts to validate models and accelerate the development of the process. This methodology, as adopted, will make it possible to apply it to any other similarly complex process such as microelectronics, new polymers, cosmetics, chemistry among others. Finally, a holistic platform providing an easy-to-use optimization pipeline in a highly variable industrial environment remains a major challenge with high added value. Such a pipeline is described in Fig.1.

## METHODOLOGY

### Data cleaning: consolidation of the model with regards to the process expertise

The objective of cleaning is to extract relevant data from raw experimental data and consolidate their format to be data science compatible. Format is variable depending on process, machines used and industry habits which must be homogenized to maintain the genericity of the pipeline. Redundant data such as duplicated data or correlated parameters are filtered out to increase the information density. Finally, multiple realization of same experiments requires specific data processing to be correctly managed by AI models. These different steps compose the proposed data cleaning pipeline for this specific application but can be easily adapted to other processes.

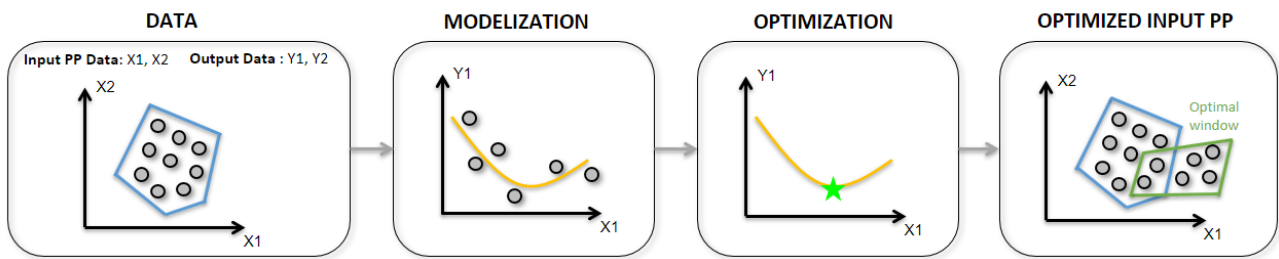

**Figure 1.** Optimization pipeline. Input Process Parameters (PP) DATA are cleaned. MODELIZATION enables digital reproduction of process past experiments. Finally, OPTIMIZATION efficiently explores the model to find the most suitable input parameters to reach targeted output parameters.

Data preprocessing consists in parsing machine files to load the data then to homogenize their content for each parameter through all the experiments. For AI algorithms, data must be consistent to extract relationships between input and output parameters. Duplicated or correlated information from the experiments is removed as they dilute relevant data. Then, based on user recommendations, rules and constraints from the process and machine such as ratios and physical limits are considered for further cleaning of the data. Special care is taken for rules and constraints as they must be conserved to reconstruct final recipes from the reduced parameters list from data cleaning.

AI based data reduction is then applied to keep only informative parameters. Two complementary approaches are proposed in our optimization pipeline. The first proposes to explore collinearity and statistics of data to rank input parameters based on their capability to explain output targets. An automatic procedure is proposed to select an optimal number of parameters, to maintain quality of the modeling and readability of the proposed experiments. Process engineers can then complement this selection by addition or removing parameters based on their knowledge. The second method relies on an exhaustive search of parameter combination based on model performances to select parameters that can be modelled accurately by AI. As the search can be long, the first approach is a good preliminary processing to optimize computing time.

Finally, the management of multiple realizations of the same experiment requires an additional step to coherent data for AI models. For the training of models, different outputs for the same inputs can lead to noise that increases the uncertainties of the predictions. Among classical methods to aggregate duplicate experiments, we can cite mean, median and dropping. Our proposition relies on median aggregation strategy to turn duplicate experiments into one. This method aims to target more realistic optimized recipes with higher confidence intervals.

## Multi-objective: extending the optimizing performances to several output criteria

The expected outcomes of a process are not unique. Multi-objective is a cutting-edge area in optimizing processes that leads to many strategies in an AI pipeline. Each section of the AI pipeline must be transformed to properly handle multiple outputs, from data cleaning to modeling and optimization. Validation and curation from the process engineers is essential to propose recipes compliant to their requirements.

A basic strategy to introduce multi-objective is to define "factors of merit" and determine merit functions that are commonly used for single processes. The use of merit functions is a fast alternative for multi-objective optimization but can lead to uncertainties as the dependencies between outputs are hidden. To overcome this major challenge, we propose adapted methods dedicated to multi-objective applications [1,2]. Data reduction algorithms are adapted to consider several outputs for filtering the most important parameters. The modeling part enables the user to choose between multiple algorithms with several outputs such as decomposition methods [3] and copulas [4,5]. Except for the specified merit functions, the developed AI pipeline considers multiple outputs and their interdependencies.

Our optimization algorithms evaluate a Pareto-front [6] of candidate solutions on which the user must set a tradeoff between multiple objectives. Pareto-front candidates are equally good solutions with variations in the importance of the different outputs. Dashboards are dynamically built to guide users in the optimization procedure and verify performances, qualities and stabilities of solutions. A customized exploration-exploitation tradeoff is accessible all along the process, with respect to a given set of optimizer hyperparameters.

## RESULTS

### Material

Experimental data from Aledia included various configurations of information to guide research support. Whereas Aledia provides obfuscated data, the company strives to actively cooperate towards comprehensive explanations about the physical process. The dataset contains 20600 input parameters and 2 output parameters.

### Data cleaning: consolidation of the model with regards to the process expertise

Mandatory cleaning is first performed: building iso-structured experiments, identifying and eliminating outliers, agreeing on parameters with real physical meaning and dealing with inter-parameter correlations. In Fig.2, the automatic data reduction led to the selection of 20 parameters, of which 5 are excluded by the expert physicists of the process and the reduced set is finally chosen. Then, deep neural networks are applied, from which a selection of models is made according to their performance in both prediction and optimization. In Fig.3, RMSE scores allow to identify potentials models according to their first sets of reduced parameters. Note that an adjusted R2 produced similar conclusion with similar variations all along the number of features.

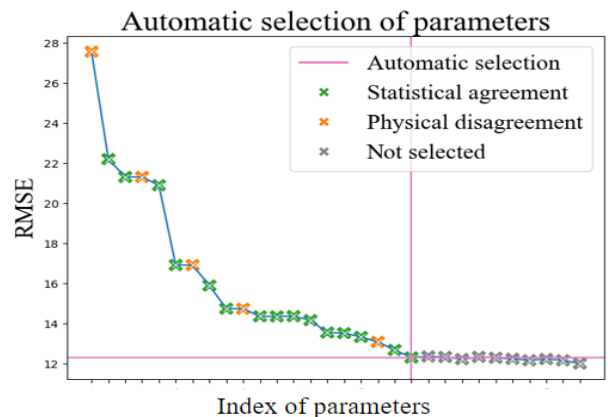

**Figure 2.** Evolution of RMSE loss all along nested models, performed on incrementally sorted number of parameters from initial set of statistical importances.

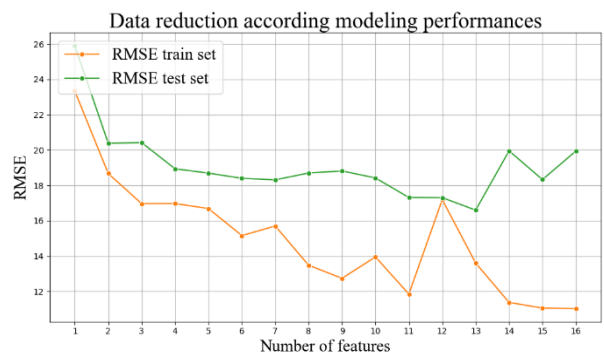

**Figure 3.** RMSE performances on iterative number of selected features, for both training and testing samples.

The models obtained are screened to eliminate overfitting and underfitting by varying optimization targets. It is also possible to select candidate models for accurate prediction inside predictable bounds. From our several models, we therefore proceed to optimize their plans to identify optimized solutions on their individual selected parameters by varying optimization targets as shown in Fig.4. From this setting of optimized parameters, complete recipes are returned. Physical and format constraints are then checked before experiments are launched by Aledia, with the most promising changes.

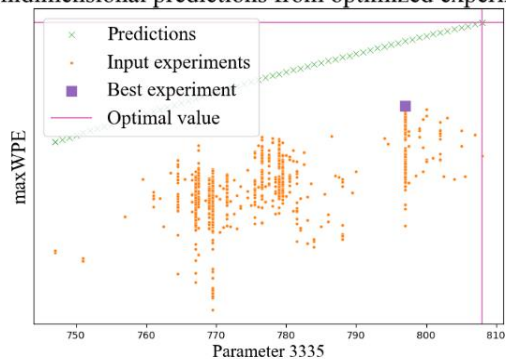

**Figure 4.** Predictions of a promising and well-fitted deep neural network on a unidimensional plan from an optimized experiment among spatial experimental range. Note that input experiments are not included in this same plan.

*Multi-objective: extending the optimizing performances to several output criteria*

Multi-objective optimization strategies especially enable the user to be part of the optimization process. A first dashboard was developed so that the user can half-open the black box by checking the performance, quality and stability of the solutions. A second dashboard enables the user to set a customized exploration-exploitation tradeoff with respect to a given set of hyperparameters of the optimizer.

All along the collaboration between the partners, opportunities to challenge this AI-protocol included various outcomes of micro-led growth. While the main focus was to predict optimized recipes by maximizing maxWPE, exploration on multi-objective targets have been performed with different outputs including V@maxWPE, Lambda peak or other morphological wires parameters such as their diameters, heights and their number. Note that some morphological parameters have been extracted by Aledia thanks to Pollen Metrology software. An example of modeling multiple output considering independent assumptions is represented in Fig.5, where ground truth is compared to predicted values for both maxWPE and Lambda peak.

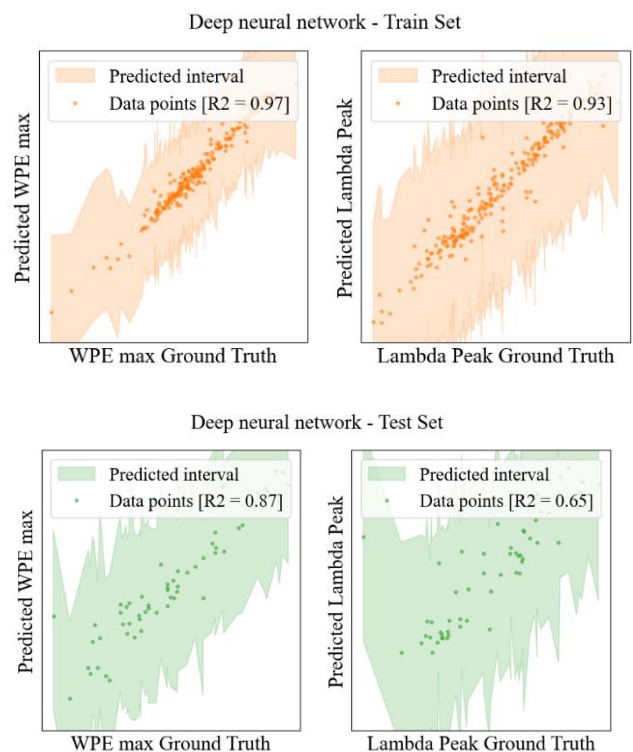

**Figure 5.** Example of modeling performances of two distinct deep neural networks under a strong assumption of independent objectives. 19 parameters as inputs enable to predict both maxWPE and Lamda peak.

Other features are available in the multi-objective optimization process. A real-time dashboard visualizes arbitrary 2D slices of the decision variables spaces, so users get an insight of the exploration of the model's landscape and of the exploitation of the best solutions found. It also evaluates the tradeoff between exploration and exploitation across the N iterations of the optimizer for a given set of the hyperparameters of the optimizer. The user can customize it for a given set of hyperparameters of the optimizer by minimizing a diversity of metrics across different runs of the optimizer. Through hyperparameter configuration, users can foster the exploration phase or the exploitation phase. The user can also check that the requested tradeoff does not entail a premature convergence to suboptimal regions of the decision variables space and does not degrade the metrics in the objective functions space.

Pareto front is the subset of the selected candidate solutions that do not have order relations and on which a tradeoff is to be done by the user. Dominated solutions are candidate solutions for which at least one solution is better in terms of the two objectives. As a result, the dominated solutions are not selected in the next $k+1^{th}$ iterations of the optimizer. Some metrics are also evaluated in real-time during the exploration and selection of the candidate solutions to assess performance, quality and stability of the Pareto fronts. A performance metrics of the $k^{th}$ Pareto front, to be maximized, evaluates the hypervolume of the dominated solutions set. A quality metric of the $k^{th}$ Pareto front evaluates the precision of its approximation via a spacing indicator between solutions. At last, a 2D Wasserstein distance evaluates the consecutive transports between the $k-1^{th}$ and the $k^{th}$ Pareto fronts so as to assess the stability of the solutions. In Fig.6, a Pareto front is represented in two-dimensional optimization based on maxWPE and Lambda peak measures.

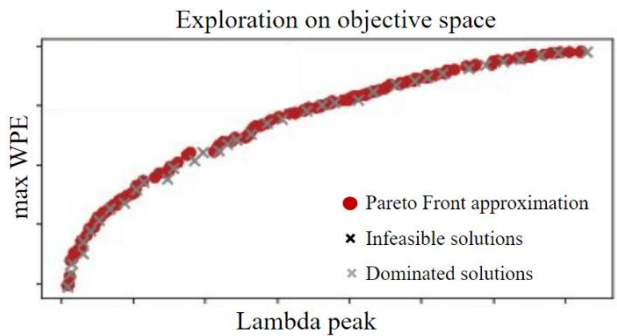

**Figure 6.** k<sup>th</sup> iteration on a dashboard displaying for {n = 1, ..., k, ..., N} iterations the exploration and selection process of candidate solutions generated by a multi-objective optimizer on a 2D objective functions space while maximizing the two targets.

## CONCLUSION

The optimization of a process using experimental raw data represents a major challenge for data science methods as data is highly variable and dense. We propose a framework for building AI models based on experimental data, with appropriate constraints and rules for effective optimization prediction.

The strategy includes automatic data processing to clean and organize data for model training, followed by a global optimization scheme. Challenges arise in balancing physical intuitions with holistic optimization and creating an automated platform accessible to non-specialists. This study revises the notion of a black box approach, emphasizing data cleaning and multi-objective criteria.

Multi-objective optimization strategies are implemented to extend performance of the optimization strategies, to explore various outcomes and allow users to impact of the optimization process. Dashboards ease user interaction and customization of exploration-exploitation trade-offs, providing insights into performance, quality, and stability of solutions.

This work opens perspectives on future research: Bayesian optimization based on surrogate models [7,8], reinforcement controls, optimal control policies and sequence based neural network on time series data. Beyond three objectives, more sophisticated optimization strategies will be adapted to higher dimensional objective spaces. All these features are available in our Smart3 software suite for process optimization

**References**

[1] Wang B, Gaussian process regression with multiple response variables. Chemometrics and Intelligent Laboratory Systems, 2020

[2] Crawshaw M, Multi-Task Learning with Deep Neural Networks: A Survey. Department of Computer Science George Mason University. 2020

[3] Jöreskog KG et al, Multivariate Analysis With Lisrel. Springer Series in Statistics; 2018

[4] Joe H, Dependence Modeling with Copulas, CRC Press, Monographs on Statistics and Applied Probability 134; 2023

[5] Yan J, Multivariate Modeling with Copulas and Engineering Applications. Handbook of Engineering statistics; 2006

[6] Goodarzi, E., Ziaei, M., & Hosseinipour, E. Z. (2014). Introduction to optimization analysis in hydrosystem engineering (Vol. 25). New York, NY: Springer International Publishing. P213.

[7] Tranter AD et al, Multiparameter optimisation of a magneto-optical trap using deep learning; 2018

[8] Pfrommer J et al, Optimisation of manufacturing process parameters using deep neural networks as surrogate model. 51st CIRP Conference on Manufacturing Systems; 2018